\documentclass[a4paper,twoside]{article}

\usepackage{epsfig}
\usepackage{subcaption}
\usepackage{calc}
\usepackage{amssymb}
\usepackage{amstext}
\usepackage{amsmath}
\usepackage{amsthm}
\usepackage{multicol}
\usepackage{pslatex}
\usepackage{apalike}
\usepackage{multirow}
\usepackage{SCITEPRESS}     

\begin{document}

\title{Binary Classification:
Counterbalancing Class Imbalance by Applying Regression Models in Combination with One-Sided Label Shifts}

\author{\authorname{Peter Bellmann, Heinke Hihn\orcidAuthor{0000-0002-3244-3661}, Daniel Braun\orcidAuthor{0000-0002-8637-6652} and Friedhelm Schwenker\orcidAuthor{0000-0001-5118-0812}}
\affiliation{Institute of Neural Information Processing, Ulm University, James-Franck-Ring, 89081 Ulm, Germany}
\email{\{peter.bellmann, heinke.hihn, daniel.braun, friedhelm.schwenker\}@uni-ulm.de}
}

\keywords{Imbalanced Classification Tasks, Binary Classification, Regression, Support Vector Machines}

\abstract{
In many real-world pattern recognition scenarios, such as in medical applications, the corresponding classification tasks can be of an imbalanced nature.  In the current study, we focus on binary, imbalanced classification tasks, i.e.~binary classification tasks in which one of the two classes is under-represented (minority class) in comparison to the other class (majority class). In the literature, many different approaches have been proposed, such as under- or oversampling, to counter class imbalance. In the current work, we introduce a novel method, which addresses the issues of class imbalance. To this end, we first transfer the binary classification task to an equivalent regression task. Subsequently, we generate a set of negative and positive target labels, such that the corresponding regression task becomes balanced, with respect to the redefined target label set. We evaluate our approach on a number of publicly available data sets in combination with Support Vector Machines. Moreover, we compare our proposed method to one of the most popular oversampling techniques (SMOTE). Based on the detailed discussion of the presented outcomes of our experimental evaluation, we provide promising ideas for future research directions.}

\onecolumn \maketitle \normalsize \setcounter{footnote}{0} \vfill

\section{\uppercase{Introduction}}\label{section:introduction}
\noindent
Imbalanced data sets, i.e.\ data sets including classes that are not represented approximately equal
\cite{DBLP:books/sp/datamining2005/Chawla05}, occur in many areas of machine learning research, such as in fraud
detection \cite{jurgovsky2018sequence}, outlier detection \cite{aggarwal2015outlier}, and
medical applications \cite{fakoor2013using}. Imbalanced data sets can lead to a
significant loss of performance in most standard classifier learning algorithms, as they
assume a balanced class distribution.
\\
\indent
In the past, a great effort has been made to address this issue \cite{sun2009classification}. One
can broadly identify the following general solution paradigms specific to class imbalance
\cite{sun2009classification}: 1) data-level approaches include different forms of
resampling \cite{xiaolong2019over} and synthetic data generation \cite{Chawla2002}
to represent the classes equally; 2) algorithm-level approaches, where the idea is to
choose an appropriate inductive bias, e.g., adaptive penalties \cite{lin2002support} and
to adjust the decision boundary \cite{wu2003class} of Support Vector Machines
\cite{vapnik2013nature}; 3) cost-level approaches adapt the misclassification costs to
reflect the imbalance; and 4) boosting approaches \cite{galar2011review}, which are common
in multi-classifier systems \cite{hihn2020specialization,Bellmann2018,DBLP:books/wi/04/K2004}.
A recently proposed example for an algorithm-level approach is the Pattern-Based Classifier for Class Imbalance Problems (PBC4cip) introduced by Loyola{-}Gonz{\'{a}}lez et al.~\cite{DBLP:journals/kbs/Loyola-Gonzalez17}. A \textit{pattern} is a set of relational statements that describe objects. A \textit{contrast pattern} is a pattern that appears significantly more often in a class than it does in the remaining classes. The idea of the PBC4cip approach is to weight the sum of support for the patterns in each class by taking the class imbalance level into account.
\\
\indent
In this work, we focus on Support Vector Machines (SVMs) and propose a novel approach, which best fits to 
the algorithm-level category, to tackle the difficulties introduced by imbalanced data sets. 
Note that classification SVMs can be reformulated as regression SVMs. Based on this property, 
we derive a label transformation technique that maps binary labels, $\{-1, +1\}$, to \emph{random} unique
positive and negative regression targets, respectively. We bound the target interval to
$[-|X^-|, |X^+|]$, where $|X^-|$ denotes the number of minority class samples and $|X^+|$ the number
of majority class samples. Moreover, we scale the targets such that the corresponding interval 
becomes symmetric. In effect, this symmetric target range alleviates some of the problems 
caused by the data sets' class imbalance, as we will show empirically.
\\
\indent
This work is structured as follows. 
In Sec.~\ref{section:relatedWork}, we shortly discuss some related work.
We introduce our approach in Sec.~\ref{section:methodology}.
In Sec.~\ref{section:dataSets}, we briefly describe the data sets. 
We present and discuss the results in Sec.~\ref{section:results}.
\section{\uppercase{Related Work}}\label{section:relatedWork}
\noindent
Most likely, data-level techniques constitute the most popular approaches to deal with class imbalance, which can be further sub-categorised in \textit{undersampling} and \textit{oversampling}. The general undersampling approach is a straightforward approach for ensemble-based classification models, i.e.~for multi-classifier systems (MCSs) which consist of a set of so-called \textit{base classifiers}. In an MCS, each base classifier is trained on a subset of the initial training set. Therefore, to overcome the issue of class imbalance, in combination with an MCS, one can train each ensemble member based on a \textit{balanced} subset of the training set, by undersampling the corresponding majority class of the current classification task.
\\
\indent
In contrast, one can also use some kind of oversampling techniques, which lead to a balanced class distribution. Thereby, one straightforward approach is the generation of additional artificial data of the minority class, for example, by simply adding noise to the initial minority class samples. Depending on the classification task, one can also apply other \textit{data augmentation} techniques, such as \textit{rotations} or \textit{shifts}, for instance, in image-based classification.
\\
\indent
One popular oversampling approach is the Synthetic Minority Over-sampling Technique (SMOTE) proposed by Chawla et al.~\cite{Chawla2002}. The SMOTE method can be briefly summarised as follows. Assume that we want to generate a new minority class data point based on the data sample $x \in \mathbb{R}^{d}, d \in \mathbb{N}$. In the first step, one has to determine the $k$ nearest neighbours to $x$, from the set of minority class samples, based on a predefined value for $k \in \mathbb{N}$, as well as on a predefined distance function. Second, one of the $k$ neighbours to $x$ is chosen randomly, for combination with $x$. Let the randomly chosen data sample from the set of nearest neighbours be denoted by $y \in \mathbb{R}^{d}$. Then, a synthetically generated new data point, $\tilde{x} \in \mathbb{R}^{d}$, is defined as follows,
\begin{equation}\notag
	\tilde{x}_{i} = x_{i} + r_{i} \cdot (x_{i} - y_{i}), \quad \forall i=1,\ldots,d,
\end{equation}
whereby $r_{i} \in [0,1]$ is a randomly generated number.
\section{\uppercase{Methodology}}\label{section:methodology}
\noindent
In the current section, we first provide the main notations. Subsequently, we briefly describe the functionality of Support Vector Machines, followed by the introduction of our proposed approach. Finally, we list the performance measures that will be used in our experimental validation.
\subsection{Formalisation}
Let $X \subset \mathbb{R}^{d}$, $d \in \mathbb{N}$, be a $d$-dimensional data set which constitutes a binary classification task. Further, let the set of labels be denoted by $\Omega=\{-1,+1\}$. In the current work, we assume that the classification task $(X,\Omega)$ is \textit{imbalanced}. Without loss of generality, we assume that the data samples from the \textit{minority} and the \textit{majority} classes are associated with the class labels $-1$ and $+1$, respectively. Moreover, by $X^{-}$ and $X^{+}$, we respectively define the set of the minority class and majority class samples, i.e.
\begin{align}
	X^{-} &:= \{ x \in X: l(x) = -1 \},
	\notag
	\\
	X^{+} &:= \{ x \in X: l(x) = +1 \},
	\notag
\end{align}
whereby $l(x)$ denotes the label of $x$.
\subsection{Support Vector Machines}
Initially, Support Vector Machines (SVMs) were introduced by Vapnik \cite{vapnik2013nature} as a classification model that is trained to separate the classes of a binary classification task. An SVM provides a unique decision boundary, which is obtained during the training phase by combining the following two objectives. First, an SVM tries to separate the provided two sample sets. Second, an SVM maximises the so-called \textit{margin}. The margin is defined as the \textit{width} between the two sample sets, based on the computed hyperplane. The data samples, which are \textit{near} to the opposing class, and which are used to define the corresponding hyperplane are denoted as \textit{support vectors}.
For two linearly inseparable classes, additional slack variables are introduced, to penalise the misclassification of training samples, based on their distance to the current hyperplane.
\\
\indent
Moreover, SVM models were modified to learn different regression tasks, e.g.~\cite{DBLP:series/acvpr/978-1-85233-929-6}. In the current work, we will denote classification SVMs and regression SVMs by cSVM and rSVM, respectively.
\subsection{Binary Regression Approach}
The main idea of our approach is based on the transformation of binary classification tasks to a specific type of regression tasks, as discussed below.
\subsubsection{Basic Idea}
As briefly discussed above, SVMs can be used as both classification and regression models. Therefore, for binary classification tasks, equivalently, one can train an rSVM in combination with the label set $\Omega=\{-1,+1\}$. Thus, to classify a (test) sample, $z \in \mathbb{R}^{d}$, one simply has to compute $sgn(rSVM(z))$, with $sgn(x)$ denoting the signum of data sample $x$.
\subsubsection{Labels Generation}
In our experiments, in Section \ref{section:results}, we will show that one can replace the label set $\Omega=\{-1,+1\}$ by a set of \textit{random} negative ($x \in X^{-}$) and positive ($x \in X^{+}$) numbers, in combination with rSVM models. Therefore, during the training phase of rSVM models, we will define the set of target labels, $\tilde{\Omega}$, as follows,
\begin{equation}
	\tilde{\Omega} := \{ -|X^{-}|, \ldots, -1, +1, \ldots, +|X^{+}| \},
\end{equation}
whereby $|\cdot|$ denotes the number of elements in the corresponding set. Thereby, we \textit{randomly} assign each $x \in X^{-}$ to one of the values from the label subset $\{ -|X^{-}|, \ldots, -1 \}$, and each $x \in X^{+}$ to one of the labels from the set $\{ 1, \ldots, |X^{+}| \}$. Note that each element of $\tilde{\Omega}$ is assigned to \textit{exactly one} data sample $x \in X$.
\subsubsection{One-Sided Labels Shift}
To overcome the issues caused by the imbalance of the given class distribution, we can further modify our proposed approach with focus on the set $\tilde{\Omega}$, and apply a one-sided \textit{labels shift} as follows.
\\
\indent
The training of an rSVM model, in combination with the set $\tilde{\Omega}$, leads to a specific function, $t: \mathbb{R}^{d} \to \mathbb{R}$. However, during the training, the target function is bounded to the set $\tilde{\Omega}$, i.e.\ to the interval $I:=\left[-|X^{-}|, +|X^{+}|\right]$, with $|X^{-}| < |X^{+}|$.
Therefore, due to the skewness of $I$, which originates from the corresponding class imbalance, $sgn(t)$ is more likely to lead to the value $+1$. To overcome this issue, we propose to balance the initial classification task by \textit{balancing the interval} $I$, i.e.\ by shifting the redefined target labels specific to the minority class to the value $-|X^{+}|$. More precisely, with $\Delta:=|X^{+}|-|X^{-}|$, by $\tilde{\Omega}_{s}$, we denote the set of shifted target labels, which we define as follows,
\begin{align}
	\tilde{\Omega}_{s} :&= \{ -(|X^{-}|+\Delta), \ldots, -(1+\Delta), +1, \ldots, +|X^{+}| \}
	\notag
	\\
	&= \{ -|X^{+}|, \ldots, -(1+\Delta), +1, \ldots, +|X^{+}| \}.
\end{align}
\subsubsection{Example \& Shift Generalisation}\label{section:generalisationShift}
Let us assume that the data set $X = \{ x_{1}, \ldots, x_{15} \} \subset \mathbb{R}^{d}$ consists of 15 data samples, with $X^{-} = \{ x_{1}, \ldots, x_{5} \}$ and $X^{+} = \{ x_{6}, \ldots, x_{15} \}$, i.e.\ $l(x_{i}) = -1, l(x_{j}) = +1$, $\forall i=1,\ldots,5$, $j=6,\ldots,15$. Then, for the training of an rSVM model, we define the set of labels as
\begin{equation}\notag
	\tilde{\Omega} := \{ -5, \ldots, -1, +1, \ldots, +10 \},
\end{equation}
since it holds, $|X^{-}|=5$ and $|X^{+}|=10$. Note that the data samples specific to the set $X^{-}$ ($X^{+}$) are randomly associated to the negative (positive) numbers from the set $\tilde{\Omega}$, such that each $x \in X$ has a unique label, i.e.\ $l(x_{i}) \neq l(x_{j})\;\forall i \neq j$. Moreover, applying the proposed labels shift leads to the following set of target labels,
\begin{equation}\notag
	\tilde{\Omega}_{s} := \{ -10, \ldots, -6, +1, \ldots, +10 \}.
\end{equation}
\indent
In our experiments, we will focus on the recognition of the minority classes. Therefore, we will analyse the performance of cSVM and rSVM models in combination with the sets $\tilde{\Omega}$ and $\tilde{\Omega}_{s}$. Moreover, we will analyse the effects of \textit{larger} shifts. More precisely, the definition of the set $\tilde{\Omega}_{s}$ can be generalised by introducing the parameter $\Delta_{m}$ which we define as follows,
\begin{equation}
	\Delta_{m} := m \cdot |X^{+}| - |X^{-}|,
\end{equation}
with a \textit{multiplier} $m$, which we will bound to the interval $[1,1.5]$, in our experiments.
We denote the corresponding set of target labels by $\tilde{\Omega}_{s}^{(m)}$, i.e.\
\begin{equation}\notag
	\tilde{\Omega}_{s}^{(m)} := \{ -(|X^{-}|+\Delta_{m}), \ldots, -(1+\Delta_{m}), 1, \ldots, |X^{+}| \}.
\end{equation}
The definition of $\tilde{\Omega}_{s}^{(m)}$ constitutes a generalisation of the set $\tilde{\Omega}_{s}$ $-$ and hence of the set $\tilde{\Omega}$ $-$ for all $m \geq |X^{-}|/|X^{+}|$. Specifically, it holds, $\tilde{\Omega}_{s} = \tilde{\Omega}_{s}^{(1)}$. Moreover, for $m=|X^{-}|/|X^{+}|$, it follows, $\Delta_{m}=0$, and therefore it holds, $\tilde{\Omega} = \tilde{\Omega}_{s}^{(m)}$.
\\
\indent
Let us recall the above example from the current section, i.e.\ $|X^{-}|=5$, and $|X^{+}|=10$. We now set the parameter $m$ to the value $2$. Then, it follows, $\Delta_{2} = 2 \cdot 10 - 5 = 15$. Hence, the corresponding shifted set of target labels is redefined as follows,
\begin{equation}\notag
	\tilde{\Omega}_{s}^{(2)} = \{ -20, \ldots, -16, +1, \ldots, +10 \}.
\end{equation}
\subsection{Performance Measures}
\begin{table}[b]
\caption{General confusion table. Rows denote true class labels. Columns denote predicted class labels. $T^{-}$: True minority. $F^{+}$: False majority. $F^{-}$: False minority. $T^{+}$: True majority. Min.: Minority. Maj.: Majority.}
\label{table:confusionGeneral}
	\centering
	\begin{tabular}{ccc}
		\hline
		$\downarrow$ true, $\rightarrow$ predicted & Min.\ Class & Maj.\ Class
		\\
		\hline
		Minority Class & $T^{-}$ & $F^{+}$
		\\
		Majority Class & $F^{-}$ & $T^{+}$
		\\
		\hline
	\end{tabular}
\end{table}
There exist different performance measures for imbalanced classification tasks that are based on the evaluation of so-called \textit{confusion matrices}, which we depict in Table \ref{table:confusionGeneral}. Using the entries of the confusion matrix, one can compute the measures \textit{specificity} ($spe$), \textit{precision} ($pre$), and \textit{recall} ($rec$), as defined in Table \ref{table:measures}.
Note that the measure recall is also denoted as \textit{sensitivity}. Using the definitions from Tables \ref{table:confusionGeneral} and \ref{table:measures}, in our experiments, we will compute the \textit{geometric mean} (G-mean) and the F1-score, which are two popular performance measures for imbalanced classification tasks, and which are defined as follows,
\begin{equation}
	\textnormal{G-mean} = \sqrt{rec \cdot spe}, 
	\quad
	\textnormal{F1-score} = \frac{2 \cdot pre \cdot rec}{pre + rec},
\end{equation}
whereby $spe$, $pre$, and $rec$ are defined in Table \ref{table:measures}.
\begin{table}[h]
\caption{Definition of different measures, based on Table \ref{table:confusionGeneral}.}
\label{table:measures}
	\centering
	\begin{tabular}{ccc}
	\hline
	Specificity ($spe$) & Precision ($pre$) & Recall ($rec$)
	\\
	$\frac{T^{+}}{F^{-}+T^{+}}$ & $\frac{T^{-}}{T^{-}+F^{-}}$ & $\frac{T^{-}}{T^{-}+F^{+}}$
	\\
	\hline
	\end{tabular}
\end{table}
\\
\indent
Note that our proposed approach does not fit completely to one of the four main counterbalancing categories from Section \ref{section:introduction}. There is no additional generation of synthetic data points, i.e.~the size of the training set is left unchanged with the initial imbalance, specific to the number of minority and majority class samples (data-level). Moreover, there is no explicit adaptation of the cost function (cost-level). Obviously, there is no correlation to any of the existing boosting-based ensembles (boosting approaches). Therefore, our proposed method fits best to the algorithm-level category. However, there is no explicit shift of the decision boundaries obtained by the corresponding SVM models. Thus, our proposed method seems to represent an additional \textit{meta-level} or \textit{target-level} category, or a specific algorithm-level subcategory.
\section{\uppercase{Data Sets}}\label{section:dataSets}
\noindent
In the current work, we will analyse the following four data sets, which are all publicly available on the UCI Machine Learning Repository\footnote{Hyperlink: http://archive.ics.uci.edu/ml} \cite{Dua:2019}.
\\
\indent
The \textbf{Arrhythmia} data set constitutes a task for the distinction between present and absent cardiac arrhythmia. Initially, this data set was defined as a $16$-class classification task, with one \textit{normal} class and $15$ classes defined as \textit{abnormal}, i.e.\ referring to some kind of cardiac arrhythmia. However, eight of the $15$ classes include less than ten samples each. Therefore, this data set is mostly used in a binary classification setting. From the initial $279$ features, we removed five features that included missing values. The provided features were extracted from different domains, which are denoted by DI, DII, DIII, AVR, AVL, AVF, as well as V1,$\ldots$,V6. Moreover, the data set also includes nominal features, such as the \textit{age}, \textit{sex}, and \textit{height}, amongst others.
\\
\indent
The UCI repository includes several data sets covering breast cancer classification tasks. In the current study, we analyse the data set that is denoted as \textit{Breast Cancer Wisconsin (Original)}, and which we will simply denote by \textbf{Breast Cancer}. This data set consists of nine ordinal-scaled features, ranging from $1$ to $10$ each. The features provided are, e.g., the \textit{clump thickness}, the \textit{uniformity of cell size}, and the \textit{uniformity of cell shape}, amongst others.
\\
\indent
The \textbf{Heart Disease} data set constitutes a binary classification task for the distinction between the presence and absence of heart diseases, based on different patients. As explained on the corresponding UCI repository database download page, initially, this data set consisted of five classes and $76$ features. However, only $13$ of the provided features are publicly available. The provided features include the patients' \textit{age}, \textit{sex}, as well as \textit{resting blood pressure}, amongst others.
\\
\indent
The \textbf{Ionosphere} data set is a binary classification task which includes radar returns from the ionosphere. The classes are denoted by \textit{good} and \textit{bad}. The first class is composed of samples that were showing evidence of some type of structure in the ionosphere. The latter class is composed of the remaining samples. All of the provided features are continuous. Technical details specific to the data set are provided in \cite{sigillito1989classification}.
\\
\indent
Table \ref{table:dataSets} summarises the properties of all data sets described above. Note that we include the Arrhythmia and Heart Disease data sets to additionally evaluate the effectiveness of our proposed method in combination with \textit{slightly} imbalanced classification tasks.
\begin{table}[h]
\caption{Data set properties. \#S: Number of samples. \#F: Number of features.
($P$\%): $P$ percent of the data belong to the minority class.}
\label{table:dataSets}
	\centering
	\begin{tabular}{lrrc}
		\hline
		Data Set & \#S & \#F & Class Distribution
		\\
		\hline
		Arrhythmia    & 452 & 274 & 207 : 245$\quad$(46\%)
		\\
		Breast Cancer & 683 &   9 & 239 : 444$\quad$(35\%)
		\\
		Heart Disease & 270 &  13 & 120 : 150$\quad$(44\%)
		\\
		Ionosphere    & 351 &  34 & 126 : 225$\quad$(36\%)
		\\
		\hline
	\end{tabular}
\end{table}
\section{\uppercase{Results \& Discussion}}\label{section:results}
\noindent
In the current section, we first provide our experimental settings, followed by the presentation of our results. Finally, we discuss the obtained outcomes, and provide some ideas for future research directions.
\begin{table*}[p]
\caption{Averaged G-mean and F1-score results in \% based on rSVMs. AR: Arrhythmia. BC: Breast Cancer. HD: Heart Disease. IO: Ionosphere. $\Omega$/$\tilde{\Omega}$/$\tilde{\Omega}_{s}^{(1.0)}$: rSVM in combination with the initial, $\Omega=\{-1,+1\}$, redefined, $\tilde{\Omega}$, redefined and shifted, $\tilde{\Omega}_{s}^{(1.0)}$, label sets (see Section \ref{section:methodology}, for the definitions). The best performing method is depicted in bold.
The improvement in performance is statistically significant, according to the two-sided Wilcoxon signed-rank test, at a significance level of 5\%, with respect to $\Omega$ and $\tilde{\Omega}$, for each data set, based on both performance measures. 
Standard deviation values are denoted by $\pm$.}
\label{table:results100_initial}
	\centering
	\begin{tabular}{lrrrrrrrr}
		\hline
		& \multicolumn{2}{c}{AR (100$\times$20 Folds)} & \multicolumn{2}{c}{BC (100$\times$20 Folds)} 
		& \multicolumn{2}{c}{HD (100$\times$10 Folds)} & \multicolumn{2}{c}{IO (100$\times$10 Folds)}
		\\
		& G-mean & F1-score & G-mean & F1-score & G-mean & F1-score & G-mean & F1-score
		\\
		\hline
		$\Omega$					 & 62.9	   	   & 58.6 		 & 94.8 	   & 93.9 		 & 79.1 	   
									 & 76.7 	   & 76.7 		 & 73.7
		\\
									 & $\pm 2.6$   & $\pm 3.1$	 & $\pm 0.9$   & $\pm 1.0$   & $\pm 2.9$
									 & $\pm 3.3$   & $\pm 2.8$   & $\pm 3.4$   
		\\
		$\tilde{\Omega}$			 & 66.5 	   & 61.8 		 & 91.6 	   & 90.7 		 & 81.5 	   
									 & 79.3 	   & 73.1 		 & 69.3
		\\
									 & $\pm 2.7$   & $\pm 3.3$	 & $\pm 1.2$   & $\pm 1.3$   & $\pm 2.4$
									 & $\pm 2.8$   & $\pm 2.5$	 & $\pm 3.1$
		\\
		$\tilde{\Omega}_{s}^{(1.0)}$ & $\bf{68.1}$ & $\bf{64.4}$ & $\bf{96.5}$ & $\bf{95.5}$ & $\bf{82.0}$ 	   
									 & $\bf{80.1}$ & $\bf{82.3}$ & $\bf{78.4}$
		\\ 
									 & $\bf{\pm 2.3}$ & $\bf{\pm 2.8}$ & $\bf{\pm 0.8}$ & $\bf{\pm 1.0}$
									 & $\bf{\pm 2.2}$ & $\bf{\pm 2.5}$ & $\bf{\pm 1.9}$ & $\bf{\pm 2.4}$
		\\
		\hline
	\end{tabular}
\end{table*}
\begin{table*}[p]
\caption{Averaged G-mean and F1-score results in \%. AR: Arrhythmia. BC: Breast Cancer. HD: Heart Disease. IO: Ionosphere. SMOTE: cSVM based on $\Omega=\{-1,+1\}$, in combination with an exactly balanced class distribution (see Section \ref{section:relatedWork}, for details). $\tilde{\Omega}_{s}^{(m)}$: rSVM in combination with the redefined and shifted label sets $\tilde{\Omega}_{s}^{(m)}$ (see Section \ref{section:methodology}, for the definitions). The best performing methods are depicted in bold.
An asterisk ($^{*}$) indicates a statistically significant difference between the best performing rSVM approach and the cSVM SMOTE method, according to the two-sided Wilcoxon signed-rank test, at a significance level of $5\%$.
Standard deviation values are denoted by $\pm$.}
\label{table:results100_extended}
	\centering
	\begin{tabular}{crrrrrrrr}
		\hline
		& \multicolumn{2}{c}{AR (100$\times$20 Folds)} & \multicolumn{2}{c}{BC (100$\times$20 Folds)} 
		& \multicolumn{2}{c}{HD (100$\times$10 Folds)} & \multicolumn{2}{c}{IO (100$\times$10 Folds)}
		\\
		& G-mean & F1-score & G-mean & F1-score & G-mean & F1-score & G-mean & F1-score
		\\
		\hline
		SMOTE 						 & 66.5 	   & 63.8 		 & 97.1 	   & 95.8 	& $^{*}\bf{83.0}$ 
									 			   & $^{*}\bf{81.2}$ & $^{*}\bf{83.0}$  & $^{*}\bf{79.2}$
		\\
									 & $\pm 2.7$      & $\pm 3.0$      & $\pm 0.7$      & $\pm 0.9$  
									 & $\bf{\pm 2.0}$ & $\bf{\pm 2.3}$ & $\bf{\pm 2.2}$ & $\bf{\pm 2.7}$
		\\
		$\tilde{\Omega}_{s}^{(1.0)}$ & 68.1 	   & 64.4 		 & 96.5 	   & 95.5 	     
									 & 82.0 	   & 80.1 	     & 82.3 	   & 78.4
		\\ 
									 & $\pm 2.3$   & $\pm 2.8$   & $\pm 0.8$   & $\pm 1.0$
									 & $\pm 2.2$   & $\pm 2.5$   & $\pm 1.9$   & $\pm 2.4$
		\\
		\hline
		$\tilde{\Omega}_{s}^{(1.1)}$ & 68.6 	   & 65.5 		 & 97.1 	   & 96.0 	     
									 & 81.9 	   & 80.1 	     & 81.6 	   & 77.2
		\\
									 & $\pm 2.0$   & $\pm 2.3$   & $\pm 0.7$   & $\pm 0.9$
									 & $\pm 2.3$   & $\pm 2.5$	 & $\pm 1.9$   & $\pm 2.4$
		\\
		$\tilde{\Omega}_{s}^{(1.2)}$ & 68.7 	   & 66.3 		 & 97.6 	   & $^{*}\bf{96.4}$ 
									 & 81.7 	   & 80.2 	     & 81.2 	   & 76.4
		\\
									 & $\pm 1.9$   & $\pm 2.1$   & $\pm 0.6$   & $\bf{\pm 0.8}$
									 & $\pm 2.5$   & $\pm 2.6$ 	 & $\pm 2.0$   & $\pm 2.4$
		\\
		$\tilde{\Omega}_{s}^{(1.3)}$ & $^{*}\bf{68.8}$  & 67.0   & $^{*}\bf{97.7}$ & $96.4$ 
									 & 81.1 	        & 79.8 	 & 81.0 		   & 76.0
		\\
									 & $\bf{\pm 2.2}$   & $\pm 2.3$   & $\bf{\pm 0.6}$   & $\pm 0.9$
									 & $\pm 2.6$   		& $\pm 2.5$   & $\pm 2.0$   	 & $\pm 2.5$
		\\
		$\tilde{\Omega}_{s}^{(1.4)}$ & $^{*}\bf{68.8}$  & 67.5 	      & $^{*}\bf{97.7}$  & 96.2 		
								     & 80.4 	        & 79.3 	      & 80.8 		     & 75.7
		\\
									 & $\bf{\pm 2.2}$   & $\pm 2.2$   & $\bf{\pm 0.6}$   & $\pm 0.9$
									 & $\pm 2.4$   		& $\pm 2.3$   & $\pm 1.9$   	 & $\pm 2.4$
		\\
		$\tilde{\Omega}_{s}^{(1.5)}$ & 68.7		   & $^{*}\bf{67.9}$  & $^{*}\bf{97.7}$  & 96.1 		 
									 & 79.7 	   & 78.9 	   		  & 80.7 		 	 & 75.4
		\\
									 & $\pm 2.2$   & $\bf{\pm 2.1}$   & $\bf{\pm 0.6}$   & $\pm 0.9$
									 & $\pm 2.4$   & $\pm 2.2$ 	 	  & $\pm 1.9$   	 & $\pm 2.4$
		\\
		\hline
	\end{tabular}
\end{table*}
\begin{table*}[p]
\caption{Averaged confusion tables. SMOTE: cSVM based on $\Omega=\{-1,+1\}$, in combination with an exactly balanced class distribution. $\tilde{\Omega}_{s}^{(m)}$: rSVM in combination with the redefined and shifted label sets $\tilde{\Omega}_{s}^{(m)}$ (see Section \ref{section:methodology}, for the definitions). Min: Minority class. Maj: Majority class. Rows denote true class labels. Columns denote predicted class labels.}
\label{table:confusion}
	\centering
	\begin{tabular}{l|rr|rr|rr|rr}
		\hline
		$\downarrow\;\;\,$ true & \multicolumn{2}{c}{Arrhythmia} & \multicolumn{2}{c}{Breast Cancer} 
		& \multicolumn{2}{c}{Heart Disease} & \multicolumn{2}{c}{Ionosphere}
		\\
		$\rightarrow$ predicted & Min & \multicolumn{1}{r}{Maj} 
								& Min & \multicolumn{1}{r}{Maj} 
								& Min & \multicolumn{1}{r}{Maj} 
								& Min & \multicolumn{1}{r}{Maj}
		\\
		\hline
		\multirow{2}{*}{SMOTE}			 			  & 132 &  75 & 232 &   7 & 97 &  23 & 95 &  31
		\\
		 											  &  74 & 171 &  13 & 431 & 22 & 128 & 18 & 207
		\\
		\hline
		\multirow{2}{*}{$\tilde{\Omega}_{s}^{(1.0)}$} & 126 &  81 & 228 &  11 & 96 &  24 & 92 &  34 	   
		\\
							 						  &  56 & 189 &  11 & 433 & 24 & 126 & 17 & 208
		\\
		\hline
		\multirow{2}{*}{$\tilde{\Omega}_{s}^{(1.1)}$} & 132 &  75 & 232 &   7 & 98 &  22 & 93 &  33 	   
		\\
							 						  &  63 & 182 &  12 & 432 & 27 & 123 & 23 & 202
		\\ 
		\hline
	\end{tabular}
\end{table*}
\subsection{Experimental Settings}
We are using the MATLAB\footnote{https://www.mathworks.com/products/matlab.html} software (version R2019b), in combination with the default parameters for linear SVM models, i.e.\ cSVMs and rSVMs with linear kernels. As evaluation approaches, we apply a $100\times20$-fold cross validation (CV) for the data sets Arrhythmia and Breast Cancer. In combination with the data sets Heart Disease and Ionosphere, we apply a $100\times10$-fold CV, due to the low number of data points. Thereby, the class distribution of each fold reflects the initial class distribution, for each of the data sets, i.e.~we apply \textit{stratified} CVs. Note that by setting the number of evaluation folds respectively to $10$ and $20$, we ensure that each of the test folds includes at least $10$ data samples from the minority class. Moreover, by applying $100$ iterations for each of the CV evaluations, we ensure a fair comparison across all analysed models.
To test for statistically significant differences between the implemented models, we will apply the two-sided Wilcoxon signed-rank test \cite{10.2307/3001968}, at a significance level of $5\%$.
\\
\indent
Note that in the current study, we focus on SVM models, since they are popular machine learning tools which are widely used in binary classification task scenarios. Moreover, we apply the default parameter settings to focus on the effectiveness of label shifts in particular.
\\
\indent
As we will discuss later in detail, in this work, we provide the basic form of our proposed balancing technique (see Section \ref{section:futureWork}). Therefore, for the comparison to the state-of-the-art, we focus on the SMOTE method, which is a very intuitive and easily interpretable, yet quite challenging approach.
\subsection{Initial Regression Experiments}
First, we focus on determining the best regression approach, based on the following three label set variants. We evaluate the rSVM performances based on the initial label set, $\Omega = \{-1, +1\}$, the modified label set, $\tilde{\Omega} = \{ -|X^{-}|, +|X^{+}| \}$, as well as the modified and \textit{symmetric} label set, $\tilde{\Omega }^{(1.0)}_{s} = \{ -|X^{+}|, \ldots, -(|X^{+}| - |X^{-}| + 1), 1, \ldots, |X^{+}| \}$, as discussed in Section \ref{section:methodology}. Table \ref{table:results100_initial} states the corresponding performance values for all of the four data sets.
\\
\indent
From Table \ref{table:results100_initial}, we can make the following observations. Based on the Arrhythmia and Heart Disease data sets, training rSVM models specific to the redefined label set, $\tilde{\Omega}$, instead of the initial label set, $\Omega$, leads to an improvement, in combination with both performance measures, G-mean and F1-score. In contrast, based on the Breast Cancer and Ionosphere data sets, training rSVM models specific to the redefined label set, $\tilde{\Omega}$, instead of the initial label set, $\Omega$, leads to a decrease of both performance measures. However, the training of rSVMs in combination with the redefined and symmetric label set, $\tilde{\Omega}^{(1.0)}_{s}$, leads to the best performance values, based on all data sets and both measures. The improvement in performance was always statistically significant, according to the two-sided Wilcoxon signed-rank test, at a $5\%$ significance level, with respect to both approaches, $\Omega$ and $\tilde{\Omega}$.
\\
\indent
However, in general, binary classification tasks are not transferred to regression tasks. Therefore, in the next step, we will compare our proposed approach to common classification SVM models.
\subsection{Classification vs.\ Regression}
In the current section, we will analyse the following two research questions. First, we will evaluate the effect of further shifting the minority class labels, such that the redefined label sets become imbalanced again, however \textit{over-representing} the minority class. Second, we will compare our proposed method to a regular cSVM model, which is trained in combination with the SMOTE method. Thereby, we apply the standard SMOTE approach in combination with the Euclidean distance, and set the number of nearest neighbours to $k=10$ (see Section \ref{section:relatedWork}). Note that we apply the SMOTE method such that the current training set is always equally distributed, i.e.\ $|X^{-}_{\textnormal{train}}|=|X^{+}_{\textnormal{train}}|$, for each validation fold, and each data set. Table \ref{table:results100_extended} states the obtained results, including the performance values for $\tilde{\Omega}^{(m)}_{s}$, with $m=1.1,1.2,1.3,1.4,1.5$.
\\
\indent
From Table \ref{table:results100_extended}, we can make the following observations. Additionally shifting the labels, with respect to $\tilde{\Omega}^{(1.0)}_{s}$, increased the performance specific to both measures, G-mean and F1-score, based on the Arrhythmia and Breast Cancer data sets. In contrast, based on the Heart Disease and Ionosphere data sets, further labels shifts, with respect to $\tilde{\Omega}^{(1.0)}_{s}$, decreased the averaged performance in general, in combination with both measures. Moreover, our proposed method outperformed the common cSVM approach, in combination with SMOTE (and the initial binary label set $\Omega$), based on the Arrhythmia, as well as Breast Cancer data set, with respect to both performance measures. On the other hand, our proposed method was outperformed by the cSVM SMOTE approach, for the other two data sets, also with respect to both performance measures.
\\
\indent
Table \ref{table:confusion} illustrates the effects of increasing the labels shift, based on the rounded confusion tables. The tables are averaged with respect to the $100$ repetitions of the $10$-fold and $20$-fold cross validations, specific to the different data sets, respectively. Comparing the cSVM SMOTE method to the $\tilde{\Omega}^{(1.0)}_{s}$ based approach, leads to the following observation, with respect to the recognition of the minority class. The cSVM SMOTE method slightly outperforms the $\tilde{\Omega}^{(1.0)}_{s}$ based approach on all of the four data sets. The effect of shifting the minority class labels is also clearly depicted in Table \ref{table:confusion}. By increasing $m$ from $1.0$ to $1.1$, we can observe that the recognition of the minority class improved, while simultaneously the recognition of the majority class decreased, based on all four data sets. The recognition of the minority class, specific to the approaches cSVM SMOTE and $\tilde{\Omega}^{(1.1)}_{s}$, is approximately the same.
\subsection{Discussion}
In the first part of our experiments, we showed that shifting the labels to a symmetric \textit{target interval} improves the performance of rSVM models, with respect to the initial class label set $\Omega$, as well as to the redefined and asymmetric target label set $\tilde{\Omega}$. Moreover, we showed that additional one-sided shifts of the labels improve the recognition of the minority class, while simultaneously decreasing the recognition rate specific to the majority class. Moreover, the performance based on our proposed approach is comparable to the performance of cSVM SMOTE models. One main advantage of our proposed method is that there is no additional generation of synthetic data. This property is especially beneficial in classification tasks where the corresponding class imbalance is \textit{relatively huge}, and the feature dimension \textit{very high}. In general, the operational cost, for determining a small set of nearest neighbours, has to be taken into account in tasks including high-dimensional data. Moreover, it is also necessary to store the additionally generated data. Applying our proposed approach avoids both of the aforementioned issues.
\\
\indent
Similar to other counterbalancing methods, it is possible to apply our proposed method also in multi-class settings, in a straightforward manner. Therefore, one can simply choose one of the existing divide-and-conquer methods, such the \textit{error-correcting output codes} \cite{DBLP:conf/aaai/DietterichB91}, including, e.g., the \textit{one-versus-one} and \textit{one-versus-all} approaches. Moreover, one can also apply our proposed method in form of \textit{cascaded classification architectures} \cite{DBLP:conf/ecml/FrankH01}, if it is possible to detect an ordinal class structure in the current classification task, as recently proposed, for instance, in \cite{DBLP:journals/access/BellmannS20,lausser2020}.
\subsection{Future Research Directions}\label{section:futureWork}
Note that in the current study, we introduced the basic form of our proposed method. Therefore, in the following, we want to provide some promising ideas for future research directions.
\\
\indent
First, one could include the parameter $m$ in the process of classifier design. As we showed in the previous section, the parameter $m$ has a strong influence on the recognition rate of the minority class. Depending on the current classification task and performance measure, one could apply some kind of optimisation techniques, e.g.~grid search, to determine \textit{optimal} values for $m$.
\\
\indent
Second, note that in the current study, we proposed to assign the labels \textit{randomly}, between the new set of labels $\tilde{\Omega}^{(m)}_{s}$ and the corresponding data samples from the majority (only positive labels) and the minority (only negative labels) classes. Therefore, we have to further analyse whether the classification performance can be improved by applying specific target label assigning approaches. To this end, one could define some kind of \textit{ordering function} $f$, $f: \mathbb{R}^{d} \to \mathbb{R}$, and apply $f$ separately to the sets $X^{-}$ and $X^{+}$. Subsequently, one can assign the target labels according to the sorted values of $f$. More precisely, $x \in X^{+}$ is assigned to the target label $|X^{+}|$, if $f(x) > f(y)$, $\forall y \in X^{+}$. Similarly, $x \in X^{+}$ is assigned to the target label $1$, if $f(x) < f(y)$, $\forall y \in X^{+}$. Analogously, based on the set $X^{-}$, the data sample specific to the lowest/highest value of $f$ is assigned to the lowest/highest negative value of the current label set $\tilde{\Omega}^{(m)}_{s}$. We assume that assigning the target labels \textit{appropriately} can lead to a \textit{smooth} regression target function, which could lead to a better classification performance. In future work, we want to evaluate different examples for the ordering function $f$, for instance,
\begin{equation}\notag
	f(x):=\sum_{i=1}^{d} | x_{i} |, \quad f(x):=\sum_{i=1}^{d} x_{i}^{n},
\end{equation}
e.g., with $n \in \{1,2\}$.
\\
\indent
Third, our proposed method is not based on any specific property of SVM models. Similar to SVMs, for instance, decision trees \cite{DBLP:books/wa/BreimanFOS84} can also be implemented as both, classification and regression models. Therefore, we aim to analyse our proposed method in combination with additional classification - or even plain regression - models. Moreover, we aim to adapt our approach to deep models, for instance, by scaling the target labels to the interval $[-1,1]$, to avoid the \textit{exploding gradient problem}.
\\
\indent
Fourth, we aim to analyse our proposed method in combination with ensemble-based classification systems. Therefore, for instance, one could design an ensemble in which each base classifier/regressor is trained on 1) different target label intervals, and/or 2) different assignments between training samples and target labels. Moreover, one could include different classification/regression models, or combine our proposed method with some approaches from the different class imbalance solution categories, which we briefly discussed in Section \ref{section:introduction}.
\\
\indent
Finally, after adapting our method specific to some or all of the modifications proposed above, we aim to provide a detailed comparison to latest state-of-the-art techniques including \textit{highly} imbalanced data sets.

\section*{\uppercase{Acknowledgements}}
\noindent 
The work of Friedhelm Schwenker and Peter Bellmann is supported by the project \textit{Multimodal recognition of affect over the course of a tutorial learning experiment} (SCHW623/7-1) funded by the German Research Foundation (DFG).
The work of Daniel Braun and Heinke Hihn is supported by the European Research Council, grant number ERC-StG-2015-ERC, Project ID: 678082, \textit{BRISC: Bounded Rationality in Sensorimotor Coordination}.
We gratefully acknowledge the support of NVIDIA Corporation with the donation of the Tesla K40 GPU used for this research.

\bibliographystyle{apalike}
{\small
\bibliography{MyCompleteLiterature}}

\end{document}